\documentclass[letterpaper, 10 pt, conference]{ieeeconf}  

\IEEEoverridecommandlockouts                              

\usepackage{graphics} 
\usepackage{epsfig} 
\usepackage{mathptmx} 
\usepackage{times} 
\usepackage{amsmath} 
\usepackage{amssymb}  
\usepackage{xcolor}
\usepackage{mathtools}
\usepackage{pdfpages}
\usepackage{graphicx}
\usepackage{algorithm}
\usepackage{algpseudocode}
\usepackage{lipsum}
\usepackage{dblfloatfix}
\usepackage{blindtext}
\usepackage{multicol,float}
\usepackage{color}
\usepackage{wrapfig}
\usepackage{tablefootnote}

\title{\LARGE \bf
Fast Shadow Detection from a Single Image Using a Patched Convolutional Neural Network*
}

\author{Sepideh Hosseinzadeh$^{1}$, Moein Shakeri$^{2}$, Hong Zhang$^{3}$
\thanks{*This paper was supported by the Natural Science and Engineering Research Council (NSERC) through the NSERC Canadian Field Robotics Network (NCFRN) and by Alberta Innovates Technology Future (AITF).}
\thanks{$^{1}$Sepideh Hosseinzadeh is with the Department of Computing Science, University of Alberta, Edmonton, Canada
        {\tt\small shossein@ualberta.ca}}%
\thanks{$^{2}$Moein Shakeri is with the Department of Computing Science, University of Alberta, Edmonton, Canada
        {\tt\small shakeri@ualberta.ca}}%
\thanks{$^{3}$Hong Zhang is with the Department of Computing Science, University of Alberta, Edmonton, Canada
        {\tt\small hzhang@ualberta.ca}}%
}

\begin{document}

\maketitle
\thispagestyle{empty}
\pagestyle{empty}

\begin{abstract}
In recent years, various shadow detection methods from a single image have been proposed and used in vision systems; however, most of them are not appropriate for the robotic applications due to the expensive time complexity. This paper introduces a fast shadow detection method using a deep learning framework, with a time cost that is appropriate for robotic applications. In our solution, we first obtain a shadow prior map with the help of multi-class support vector machine using statistical features. Then, we use a semantic-aware
patch-level Convolutional Neural Network that efficiently trains on shadow examples by combining the original image and the shadow prior map. Experiments on benchmark datasets demonstrate the proposed method significantly decreases the time complexity of shadow detection, by one or two orders of magnitude compared with state-of-the-art methods, without losing accuracy.  

\end{abstract}

\section{INTRODUCTION}

Dealing with shadows is one of the most fundamental issues in image processing, computer vision and robotics. Shadows are omnipresent in outdoor applications and must be taken into account in the solutions to standard computer vision and robotics problems such as image segmentation~\cite{segment}, change detection~\cite{shakeri_ICCV}, place recognition~\cite{place_rec}, background subtraction~\cite{shakeri_small,shakeri_corola}, visual robot localization~\cite{Corke,shakeri_dynamic} and navigation~\cite{Maddern}. Unfortunately in most of these cases images are strongly influenced by shadow at different times of a day, making it difficult to interpret or understand a scene. Although in some applications people have attempted to address this challenge by relying on robust features with impressive results, these features often do not provide sufficient invariance to shadow.

The problem of detecting shadow is a well-studied research topic, and many methods have been proposed~\cite{Corke, Maddern, Finlayson, Jiang, Shakeri, Barrow, Zhu, Khan, Khan2, Vincente_rev}. Existing methods can be categorized into two major groups. The first group of methods alleviate or remove the effect of shadow by providing an invariant representation of the image~\cite{Corke,Maddern,Finlayson,Finlayson2,Shakeri, Barrow}. Most of these methods model the process of image formation to build shadow-free images. Although these methods are effective to some extent, all of them have the limitation in terms of dealing with non-Plankian source of light, narrow-band color camera and environment calibration. They also tend to lose information in the shadow-free representation that can be important for scene understanding. The other group of methods to deal with shadow rely on a learning framework~\cite{Gue,Vicente,Zhu, Khan, Khan2, Vincente_rev}. These methods specifically focus on shadow detection while attempting to keep the original color and intensity of images. Unfortunately, these methods still have difficulty in robotics applications that we will elaborate in the next section. 

State-of-the-art shadow detection methods come from the second group above and are based on convolutional neural networks (CNN). In this paper we present a novel and fast method, also based on CNN. Our method detects shadows of an image without any assumptions about image quality or the camera and it is therefore  appropriate for the robotics applications. To develop an efficient shadow-detection algorithm, rather than labeling individual pixels, we work with super-pixels, obtained through segmentation. Subsequently, we extract color and texture features from each super-pixel and, with the help of a trained SVM, compute a shadow prior in terms of the probability of a super-pixel being shadow. Then, we use the combination of the original image and the obtained shadow prior as the input to a patch-level CNN to compute the improved shadow probability map of the image. The edge pixels between the super-pixels are further refined by running the same patch-level CNN a second time, to produce the final shadow detection result. We will show that the proposed method can provide comparable results with existing deep shadow detection methods due to the use of the combination of texture features and deep neural networks, but works much faster in both training and detection phases than existing CNN based methods, due to the use of super-pixels. This method enables us to detect shadow in an image in robotic applications, a task that was not possible before due to the high time complexity.

The remainder of this paper is organized as follows. In Sections~\ref{related}, we discuss related works in shadow detection in further detail. In Section~\ref{proposed}, we introduce our novel method to this problem, focusing on improving the efficiency of an existing deep neural network based solution. Comparative experimental results on benchmark datasets are described in Section~\ref{results}, and finally Section~\ref{conclusion} summarizes our method and concludes the
paper.

\begin{figure*}[t]
\includegraphics[width=\linewidth]{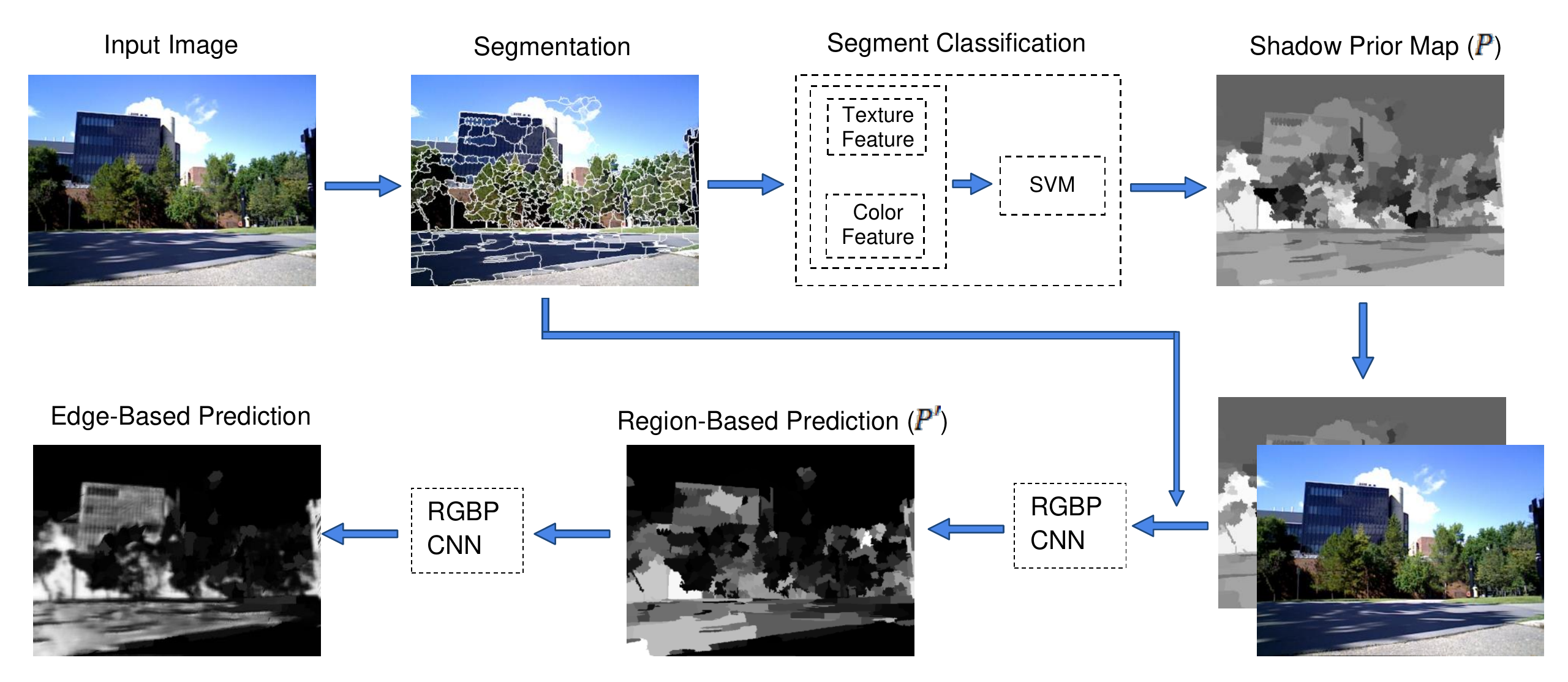}
\centering
\caption{Proposed method pipeline. For obtaining shadow prior map, input image is segmented by mean shift algorithm, then for each segment, we obtain the confidence of being shadow using SVM. Color and texture features are input of SVM. This shadow prior (P) is attached to the RGB image for using RGBP patched-CNN. In detection of shadow algorithm, we have two steps region and edge based predictions to obtain the final shadow probability map. The segmentation information obtained by mean shift is used in region based prediction step.}
\label{sample_fig}
\end{figure*}

\section{RELATED WORKS}
\label{related}

The importantce of detecting shadow in a single image has been well investigated in computer vision and robotics community. One common approach in robotics community computes ``intrinsic images"~\cite{tanenbaum} by decomposing an image into its reflectance and illuminance constituents~\cite{Corke,Shakeri}. As discussed in the previous section, these methods have restrictive assumptions on illumination and sensor. To relax these assumptions, data-driven methods have been proposed, which work with images in grayscale or color space based on a learning framework to learn shadow in different situations~\cite{Lalonde,Gue,Vicente,Zhu}. Zhu {\it{et al.}}~\cite{Zhu} proposed a method that classifies regions based on statistics of intensity, gradient, and texture, computed over local neighborhoods, and refines shadow labels by exploiting spatial continuity within a conditional random field (CRF) framework~\cite{CRF}. Lalonde {\it{et al.}}~\cite{Lalonde} find shadow boundaries by comparing the color and texture of neighboring regions and employing a CRF to encourage boundary continuity. To benefit from global information, Guo {\it{et al.}}~\cite{Gue} proposed a region based method, which can model long-range interaction between pairs of regions of the same material, with two types of pairwise classifiers, under similar/different illumination conditions. Then, they incorporated the pairwise classifier and a shadow
region classifier into a conditional random field (CRF) via graph-cut optimization~\cite{Graph_cut}. Vicente {\it{et al.}}~\cite{Vicente} proposed a multi-kernel model to train a shadow support vector machine (SVM). Their multi-kernel model is a summation of base kernels, one for each type of local features. The main limitation of this model is the assumption of equal importance for all features. To avoid this assumption, they proposed leave-one-out kernel optimization~\cite{Vincente_rev} in which the parameters of the kernel and the classifier are jointly learned. They also used least square SVM (LSSVM), which has a closed form solution and therefore is faster than SVM. However, their approach is still computationally expensive.
Although these methods reviewed above provide good accuracy in some cases, they run on the order of many minutes or seconds, and are not applicable in robotics due to this high time complexity.

Recently, some end-to-end deep learning frameworks have been proposed to learn the most relevant features for shadow detection. They outperform the state-of-the-art methods that use hand crafted features. The first method in this category is proposed by Khan {\it{et al.}}~\cite{Khan, Khan2}. They train two CNNs, one for detecting shadow regions and the other for detecting shadow boundaries. They also train a unary classifier by combining the two CNNs, and the per-pixel predictions are then fed to a CRF for enforcing local consistency.~\cite{Shen} proposes a structured deep edge detection for shadows and shows that using structured label information, local consistency over pixel labels can be improved. More recently, Vicente {\it{et al.}}~\cite{Vicente2} propose a method by combining an image level fully connected network (FCN) and a patch-based CNN. They train the FCN for semantically aware shadow prediction and use the output of the FCN as shadow prior with the corresponding input RGB image to train the patched-CNN from a random initialization. This method produces excellent result but is unsatisfactory in terms of its time complexity.

In this paper, we propose a novel method based on deep learning with a shadow prior. Like~\cite{Shen}, our method can detect shadow from a single image, but at a much lower time complexity than all existing methods based on deep learning.  The key insight of our method is that it performs shadow detection on a per super-pixel basis. Initial result from such a detection method would produce boundary effects between super-pixels.  We overcome such artifacts with a post-processing step using edge refinement. 



 
\section{PROPOSED METHOD}
\label{proposed}
In this section we describe our proposed method to detect shadows from a single image. Our method uses two steps to learn shadow from training images. First we obtain a prior map that we call shadow prior using a trained SVM on color and texture features, and then we train a patched-CNN using the original images and their shadow priors obtained from the first step. These two steps will be detailed in Section~\ref{sh_prior} and~\ref{patched_cnn}, respectively. For the detection of shadows in a given image, we compute its shadow prior  first with the SVM and use the prior and the image as input to the trained patched-CNN, considering only the center pixel of each super-pixel of the input as the representative in order to reduce the computational time. In doing so, however, the super-pixels near object and shadow boundaries tend to produce unreliable ``edge effects". To overcome this problem, we refine prediction labels for the edge pixels along super-pixel boundaries, using the patch-CNN again. In other words, algorithm makes use of the trained patched-CNN twice. The output of the second patched-CNN provides the final detected shadow areas. The edge refinement step of our algorithm will be discussed in Section~\ref{detection}. 

\subsection{Computing Shadow Prior}
\label{sh_prior}
Shadow prior computation involves steps shown in the first row of Fig.~\ref{sample_fig}. We first segment the image using mean shift algorithm~\cite{mean_shift} to obtain superpixels. Second, using a trained classifier on texture and color features, we estimate the shadow probability of each region. Segmentation enables us to estimate the shadow probability for each region instead of each pixel and therefore, the computation time is significantly decreased.

In general, a shadowed region is darker and less textured than a non-shadow region~\cite{Zhu}. Therefore, the color and texture of a region can help to predict whether it is in shadow. We exploit this observation and represent color with a histogram in L*a*b space, with 21 bins per channel, as was successfully applied in~\cite{Gue}. We represent texture with the texton histogram provided by~\cite{Martin}. We train our SVM classifier with the color and texture features with a $\chi^2$ kernel and slack parameter C = 1~\cite{libsvm}. We define the shadow prior of each super-pixel, as the log likelihood output of this trained classifier. In the subsequent step, we use this shadow prior as a critical input to our patched-CNN. 

\subsection{Training Patched-CNN with the Shadow Prior}
\label{patched_cnn}
In the next step of our shadow detection pipeline, we employ a patch-wise CNN to predict shadow. Current research in~\cite{Vicente2} showed that using patches of an image with a specific size has two benefits in the case of shadow recognition. First, these patches include enough local pattern of the image and more global information in a large-range of neighborhood pixels than pixel-based methods. Secondly, using patches we are able to provide more training samples with different patterns from a limited number of labeled images. As discussed, one of the challenging problems in the case of shadow detection is the number of training samples, which can significantly affect the accuracy of a deep neural network. Unfortunately, available shadow benchmark datasets are small due to the high cost of shadow annotation. This patch-wise structure enables us to provide a huge number of patches of shadow and non-shadow areas for training that can increase the overall accuracy of the network.

We utilize the network architecture used in \cite{Vicente2}. The deep network has six convolutional layers, two pooling layers, and one fully connected layer. The input of this network is a 32$\times$32 RGBP patch selected from combining $RGB$ image and shadow prior image $P$. The output is the shadow probability map of the patch.
We select equal number of patches for training in three classes as follows.

\begin{itemize}
\item {\textbf{Shadow patches:}} Since we are going to learn shadow areas, we first select patches from shadow regions. 

\item {\textbf{Non-shadow patches:}} We select patches from non-shadow image locations randomly to include patches of various textures and colors. Also, these selected patches prevent overfitting. 

\item {\textbf{Shadow-Edge patches:}} We also select patches on edges between shadow and non-shadow regions, to learn the shadow boundaries. Since the ground-truth is binary, locations of all shadow edges can be extracted accurately.
\end{itemize}

Using this strategy, we are able to provide millions of patches from thousands of images. The loss function of the network is the average negative log-likelihood of the prediction of every pixel.

\begin{figure*}[b!]
    \centering
    \includegraphics[width=\linewidth, height=0.37\linewidth]{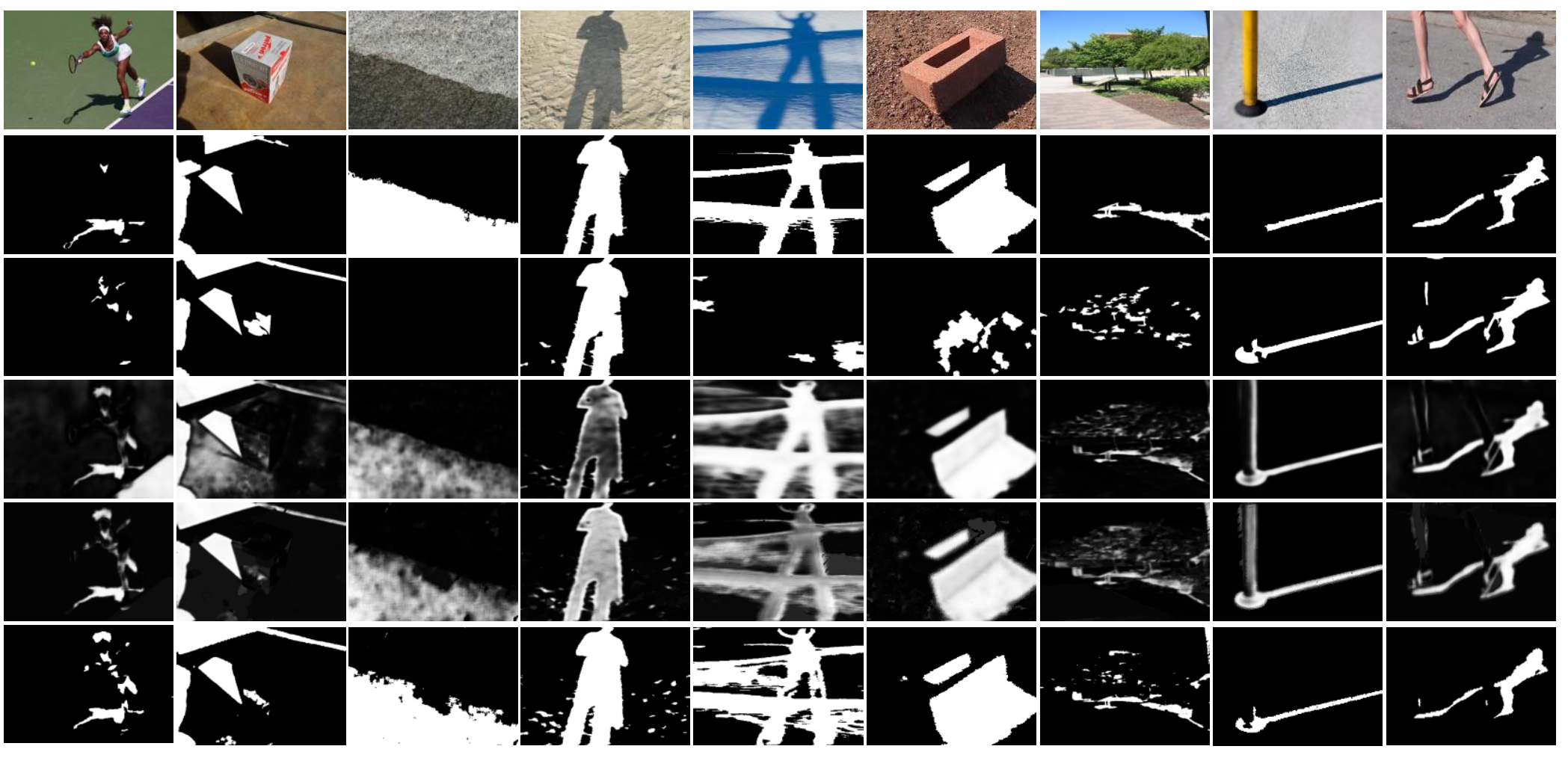}
    \caption{Comparison of our qualitative results with the results of other methods. Rows from top to bottom: input images, ground truths, results of unary-pairwise method, results of stacked-CNN, obtained probanility map of our method, binary mask of shadows based on the probability map of our method.}
    \label{fig:qual_res}
\end{figure*}

\subsection{Edge Refinement of Super-Pixel Labels}
\label{detection}
For the detection of shadows in a given image, only the patches centered in the super-pixel center are used and the average value of each predicted patch is assigned to all pixels of the super-pixel. These predictions made by the patched-CNN are local, and the prediction results near shadow boundaries are poor. To improve the accuracy of our detection algorithm, higher level interaction between the regions is needed. Therefore, in this final step we process the edge pixels between the regions by patched-CNN once again, shown in the bottom row of Fig.~\ref{sample_fig}. We only process those pixels that are on edges between the segments, labeled as $R(S)$ and defined as:

\begin{equation}
R(S) = \{s_i \in S| s_i \geq \alpha \max_{1 \leq i \leq m}(s_i) \}
\label{eq:thresh_test}
\end{equation}
$R(S)$ contains those segments with the higher probability than a threshold of maximum shadow probability in the region-based prediction $P'$. $\alpha$ is a constant threshold (equal to 0.2 in our implementation) and $m$ is the number of superpixels or regions in the image or the region-based prediction $P'$. Absolute non-shadow regions always provide a very low shadow probability in the shadow prior map, and (\ref{eq:thresh_test}) only filters out those regions. This thresholding step will reduce the number of pixels to be refined and the total time of this step.

For each boundary pixel $(x, y)$ between segments that is included in $R(S)$, a window patch with size $32 \times 32$ surrounding the pixel $(x, y)$ from its shadow prior and corresponding original image are given to the patched-CNN to predict the shadow probability for that patch. Then we set the edge pixel $(x, y)$ and its 8 neighbor pixels' probability values to be average probability value of these 9 pixels. 

This step can integrate the segmented probability maps obtained in previous step and the final shadow probability map becomes smooth.

\section{EXPERIMENTAL RESULTS}
\label{results}
In this section we perform a set of experiments to evaluate our proposed method and compare it with other state-of-the-art methods. We first use three challenging available datasets ``UCF"~\cite{Zhu}, ``UIUC"~\cite{Gue}, and ``SBU"~\cite{Vicente2} for shadow detection to evaluate quantitatively the proposed method. The number of images in each of dataset in our experiments is as follows.
\begin{itemize}
\item{\textbf{UCF Dataset}}: This dataset contains 355 images with manually labeled pixel-based ground  truth.\\
\item{\textbf{UIUC Dataset}}: This dataset contains 108 images (32 train images and 76 test images) with pixel-based ground truth.\\
\item{\textbf{SBU Dataset}}: This new dataset contains 4,727 images (4,089 train images and 638 test images) with pixel-based ground truth.\\
\item{\textbf{Combined Dataset}}: Both UCF and UIUC datasets include an insufficient number of images, and to evaluate the propose method we need to select a portion of these datasets as training samples. Since our proposed method works on patches, we are able to create many patches from the datasets for the training phase. However, to be fair in comparison with other methods, we combine UCF, UIUC, and SBU datasets to train for all methods. The combined dataset includes 5,078 images. We randomly selected 25\% of the images for testing, and the rest for training. 
\end{itemize}

In addition, we use two other datasets ``UACampus" and ``St. Lucia"~\cite{Lucia} to obtain qualitative results of detecting shadows on roads, one of the common problems in robot applications.

\subsection{Evaluation Metrics}
\label{evaluation}
To evaluate the proposed method in terms of detection accuracy, we use three evaluation metrics as follows.
\begin{equation}
Shadow Accuracy =\frac{TP}{all\,\, shadow\,\, pixels}
\end{equation}
\begin{equation}
Non-shadow Accuracy = \frac{TN}{all\,\, non-shadow\,\,pixels}
\end{equation}
\begin{equation}
Total Accuracy = \frac{TP+TN}{all\,\,pixels}
\end{equation}

For comparison in terms of computational efficiency, we simply use the execution time of shadow detection in a single image as the performance metric. Therefore, a total of four performance metrics, three for accuracy and one for efficiency, are considered in our experimental evaluation.

\subsection{Results on Benchmark Datasets}
In this section we evaluate our proposed method and compare it with Stacked-CNN~\cite{Vicente2} and Unary-Pairwise~\cite{Gue}. We select Stacked-CNN as a recent shadow detection method based on deep learning framework that uses a shadow prior map. We choose the unary-pairwise method since it is one of the best statistical methods to detect shadows from a single image. Fig.~\ref{fig:qual_res} shows example results of these methods and ours. 

In the third row, although the unary-pairwise method provides acceptable results in some cases, it completely fails in the first, third, fifth, and sixth columns. The fourth row of Fig.~\ref{fig:qual_res} shows the results of Stacked-CNN, which in the most cases are comparable with our method (the fifth row). The last row shows the binary shadow mask of the proposed method by a constant threshold.




Table~\ref{tab:sbu_res} shows the quantitative results on SBU dataset. The values shown are the average of the performance metrics on all test images. Although the total accuracy of the proposed method is not the best, with respect to shadow accuracy, our method outperforms the other methods. The goal of the proposed method is providing a fast shadow detection method without losing the accuracy. Thereore, in Table~\ref{tab:sbu_time_gpu} we show the execution time of training and testing phases of the proposed method. Results in both Tables~\ref{tab:sbu_res} and~\ref{tab:sbu_time_gpu} illustrate that the proposed method works an order of magnitude faster than the statistical methods and two orders of magnitude faster than the deep learning competing method with the comparable accuracy. This is almost an entirely a result of predicting shadow/non-shadow labels on super-pixels rather than pixels, even with the additional cost to pay for refining the boundary pixels.

\begin {table}[t]
\caption[Evaluation of shadow detection methods on SBU dataset]{Evaluation of shadow detection methods on SBU dataset} \label{tab:sbu_res}
\begin{center}
\begin{tabular}{ |p{1.77cm}||p{1.48cm}|p{1.48cm}|p{1.55cm}|  }
 \hline
 Method& Accuracy/std& Sh-Acc/std& Non-sh-Acc/std\\
 \hline
 Stacked-CNN & \textbf{0.8850 / 0.13} & 0.8609 / 0.23 &  0.9059 / 0.15\\
 Unary-Pairwise &   0.8639 / 0.14  & 0.5636 / 0.35  & \textbf{0.9357 / 0.12}\\
 Our method & 0.8664 / 0.14 & \textbf{0.8987 / 0.20} &  0.8773 / 0.15\\
 \hline
\end {tabular}
\end{center}
\end {table}
\begin {table}[t]
\caption[Time complexity of shadow detection methods on SBU dataset, using GPU]{Time complexity of shadow detection methods on SBU dataset, using GPU} \label{tab:sbu_time_gpu}
\begin{center}
\begin{tabular}{ |p{1.77cm}||p{1.48cm}|p{1.48cm}|p{1.55cm}|   }
 \hline
Method& Testing (hours)& Testing (sec/image) & Training (hours)\\
 \hline
 Stacked-CNN    & 25.08 & 141.56 & 9.4+TrFCN\\
 Unary-Pairwise&   9.13 & 51.56& -\\
 Our method & \textbf{0.33} & \textbf{1.87}& \textbf{3.1}\\
 \hline
\end {tabular}
\end{center}
\end {table}
\begin {table}[b]
\caption[Evaluation of shadow detection methods on combined dataset]{Evaluation of shadow detection methods on combined dataset} \label{tab:combined_res}
\begin{center}
\begin{tabular}{ |p{1.77cm}||p{1.48cm}|p{1.48cm}|p{1.55cm}|  }
 \hline
 Method& Accuracy/std& Sh-Acc/std& Non-sh-Acc/std\\
 \hline
 Stacked-CNN  & 0.9044 / 0.12 & \textbf{0.8614 / 0.18} &  0.9140 / 0.13\\
 Unary-Pairwise&   0.8835 / 0.13  & 0.6374 / 0.32  & \textbf{0.9366 / 0.11}\\
 Our method & \textbf{0.9103 / 0.11} & 0.8527 / 0.20 &  0.9248 / \textbf{0.11}\\
 \hline
\end {tabular}
\end{center}
\end {table}
\begin {table}[b]
\caption[Time complexity of shadow detection methods on combined dataset, using GPU]{Time complexity of shadow detection methods on combined dataset, using GPU} \label{tab:combined_time_gpu}
\begin{center}
\begin{tabular}{ |p{1.9cm}||p{1.48cm}|p{1.48cm}|p{1.55cm}|  }
 \hline
 Method& Testing (hours)& Testing (sec/image) & Training (hours)\\
 \hline
 Stacked-CNN & 27.77 & 78.75 & 9.08+TrFCN \\
 Unary-Pairwise 
 &   20.77 & 58.87 & -\\
 Our method & \textbf{0.55}& \textbf{1.55}& \textbf{3.05}\\
 \hline
\end {tabular}
\end{center}
\end {table}

We also evaluate the proposed method and compare it with the other methods on the combined dataset to investigate the effect of increasing the number of images for training and testing. Table~\ref{tab:combined_res} shows that our method is still comparable with other methods in terms of accuracy, and Table~\ref{tab:combined_time_gpu} shows that the execution time of our method is significantly less than the other two methods, as was the case on individual datasets.

\subsection{Shadow Detection in Road Detection Application}
\label{road_res}
\begin{figure}
    \centering
    \includegraphics[width=\linewidth, height=0.5\linewidth]{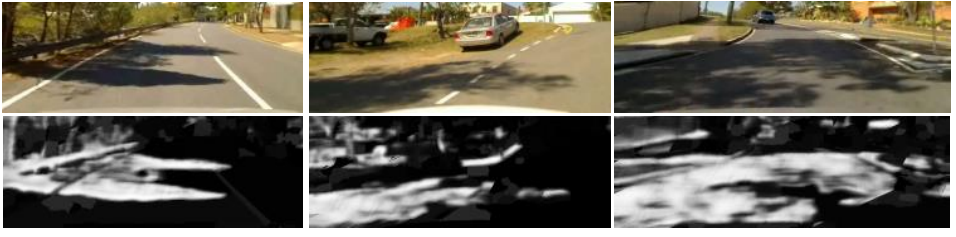}
    \caption{Qualitative results of the proposed method on smaple images of the St.Lucia dataset.}
    \label{fig:qual_road1}
\end{figure}

\begin{figure}
    \centering
    \includegraphics[width=\linewidth, height=0.5\linewidth]{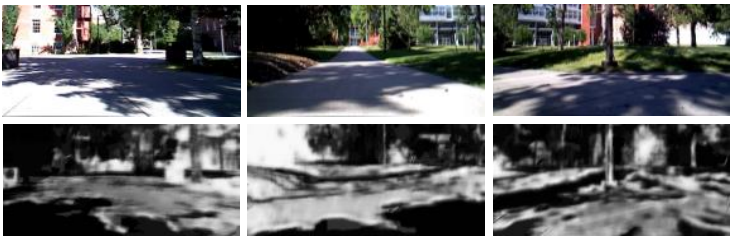}
    \caption{Qualitative results of the proposed method on smaple images of the UACampus dataset.}
    \label{fig:qual_road2}
\end{figure}

To illustrate the utility of our method in a real application, we consider the problem of detecting shadow on roads in this section. Cast shadows on roads can cause difficulty or mistakes in the scene interpretation or segmentation for this application. To determine the performance of our method in this application, we apply the proposed method on St. Lucia and UACampus datasets to show the potential of the method in detecting shadows on roads. Figs.~\ref{fig:qual_road1} and~\ref{fig:qual_road2} show the results of our method in terms of shadow probability maps of sample images.  It is clear from these examples that our method is able to generate a probability map of high accuracy, and can serve as a useful building block for road detection in, for example, autonomous driving. 

\begin{figure}[h]
    \centering
    \includegraphics[width=\linewidth, height=0.5\linewidth]{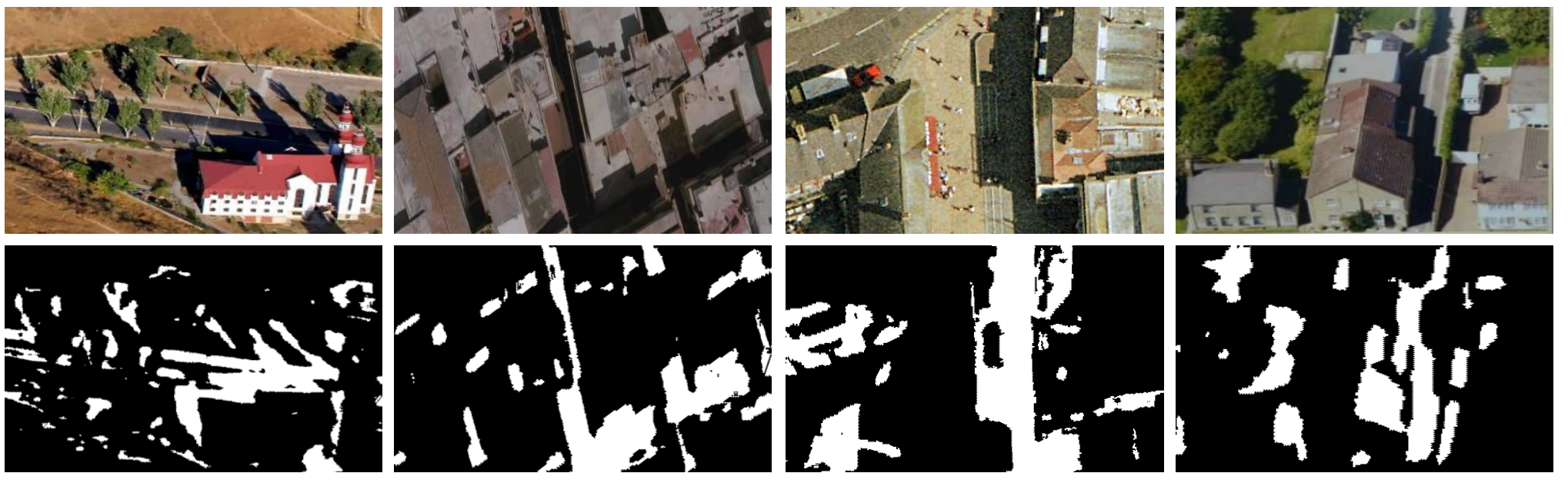}
    \caption{Qualitative results of the proposed method on aerial images. Last row shows the binary mask of the shadow probability map.}
    \label{fig:aerial}
\end{figure}

We also apply the proposed method on aerial images to detect shadows, a common problem for many applications that rely on the aerial images. Fig.~\ref{fig:aerial} shows the qualitative results of the proposed method to detect shadows in the aerial images. Once again, the results are highly accurate, and our method can directly contribute to solutions in aerial imaging applications.

\section{CONCLUSION}
\label{conclusion}

In this paper, we have presented a method for accurately detecting shadow in a single image.  Our method combines  traditional color and texture features and deep learning in a novel way, and achieves start-of-the-art performance in terms of detection accuracy and out-performance state-of-the-art in terms of computational efficiency. Our method uses color and texture features to compute a shadow prior map by training an SVM. The prior map and the original input image are then used as input to a patched-CNN to compute shadow probability map, one for each super-pixel, to achieve the desired computational efficiency. In the final step, we refine the prediction result of the patched-CNN by re-estimating the class labels of boundary pixels between super-pixels with the same patched-CNN. Extensive experimental results demonstrate that the proposed method works significantly faster than the existing deep or statistical methods, by one to two orders of magnitude, without losing the accuracy. 





\end{document}